\documentclass{esannV2}
\usepackage[dvips]{graphicx}
\graphicspath{ {./images/} }
\usepackage[latin1]{inputenc}
\usepackage{amssymb,amsmath,array}

\begin{document}
\title{3D U-Net for Segmentation of Plant Root MRI Images in Super-Resolution}

\author{Yi Zhao$^1$, Nils Wandel$^1$, Magdalena Landl$^2$, Andrea Schnepf$^2$ and Sven Behnke$^1$\thanks{This research was supported by grants BE 2556/15 and SCHN 1361/3-1 of German Research Foundation (DFG).}
%
\vspace{.3cm}\\
%
1 -- University of Bonn, Computer Science Institute VI, 53115 Bonn, Germany\\
2 -- FZ J\"ulich GmbH, Institute of Bio- and Geosciences 3, 52425 J\"ulich, Germany
}

\maketitle

\begin{abstract}
Magnetic resonance imaging (MRI) enables plant scientists to non-invasively study root system development and root-soil interaction. Challenging recording conditions, such as low resolution and a high level of noise hamper the performance of traditional root extraction algorithms, though. 
We propose to increase signal-to-noise ratio and resolution by segmenting the scanned volumes into root and soil in super-resolution using a 3D U-Net. Tests on real data show that the trained network is capable to detect most roots successfully and even finds roots that were missed by human annotators. 
Our experiments show that the segmentation performance can be further improved with modifications of the loss function.

\end{abstract}

\section{Introduction}

\begin{figure}[h!]
\centering
\includegraphics[scale=0.6,width=315pt]{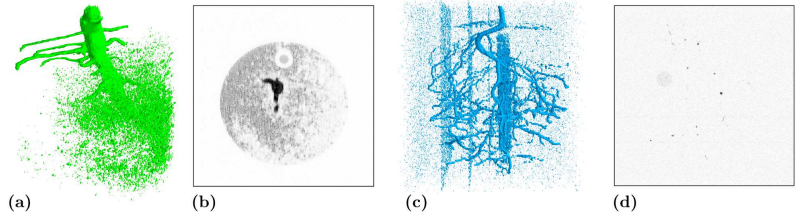}
\caption{3D visualizations of noisy root images and original MRI image slices. (a)(c) each depicts the 3D visualization of a 3D MRI image, and (b)(d) each shows a horizontal slice of the 3D image on its left.}\label{fig:mri_and_gt}
\end{figure}

In order to gain a better understanding of plant growth and its response to environmental influences such as droughts, sinking groundwater levels or climate change, plant scientists investigate root-soil interaction processes. Magnetic resonance imaging (MRI) non-invasively measures both roots and soil~\cite{stingaciu2013situ}. Low resolution of MRI scans---compared to the diameter of thin roots---and high noise (Figure~\ref{fig:mri_and_gt}) due to ferromagnetic particles or low contrast between root and soil water signal hamper the application of traditional root extraction algorithms~\cite{schulz2013plant}. For these reasons, root reconstruction is often done manually. To enable automatic reconstruction, the signal-to-noise ratio (SNR) and resolution of these MRI scans need to be enhanced.

In this work, we propose using a 3D U-Net~\cite{cciccek20163d} to segment scans of pots containing plant roots into root and soil in super-resolution. As training data is scarce, we create synthetic MRIs by rendering roots in 3D and combine them with real MRI scans of pure soil. In this way, we improve segmentation performance in soil regions, compared to models that were trained on synthetic data only~\cite{uzman2019learning,horn2018superresolution}. 
Furthermore, we investigated the influence of different loss modifications. By doing this, the segmentation performance gets improved, especially for thin roots which are difficult to detect intactly.





\section{Related Work}
Recently, CNNs have achieved remarkable performance in 3D image segmentation. Havaei et al.~\cite{havaei2017brain} use 2D CNNs to segment horizontal slices of 3D images and combine them into 3D outputs. However, 2D CNNs can only make limited use of spatial information along the vertical dimension. 3D CNNs can overcome this issue  by learning 3D features from input images~\cite{cciccek20163d}.

Since their introduction in 2016, 3D U-Net~\cite{cciccek20163d} has become a commonly used architecture for segmentation tasks of 3D data.
The structure of 3D U-Net consists of a downsampling encoder and an upsampling decoder. The encoder extracts increasingly abstract features from an increasingly broader context, while the decoder restores the resolution from the downsampled feature maps. Moreover, to share the high-resolution features extracted in the encoder with the decoder, shortcut connections are established between them. 

For image super-resolution, deep neural networks show state-of-the-art performance. 
To achieve a higher resolution in the network output, the input image needs to be upsampled. This can be done either by upsampling the input using interpolation~\cite{dong2014learning}, or within the network using methods like transposed convolution~\cite{dong2016accelerating}. The latter results in better performance because the interpolation is learned directly from the data.

Root extraction algorithms were developed to automatically extract structural models of roots from 3D MRI images~\cite{schulz2013plant}. However, their performances degrade significantly in the presence of low SNR or low resolution~\cite{schulz2013plant}. Uzman et al.~\cite{uzman2019learning} increased the SNR and resolution of MRI images using a 2D RefineNet.
Due to the limited number of real images, synthetic MRI data was used during training. While the RefineNet is able to detect most roots, it also showed a non-negligible amount of false positives in some cases. 
Horn et al.~\cite{horn2018superresolution} tried to complete the same task with a 3D CNN. However, because of memory constraints of the graphics processing unit (GPU), the network structure was relatively shallow. This resulted in fewer false positives, but at the same time significantly more false negatives for thin roots, leading to large gaps between detected root fragments.

\section{Segmentation in Super-Resolution}
We segment the MRI scans in double resolution, i.e. the task is to map an input image $I\in \mathbb{R}^{x\times{y}\times{z}}$ to a binary segmentation output $S\in \mathbb{B}^{2x\times{2y}\times{2z}}$. 
To compute the mapping, a 3D U-Net architecture was chosen. It is relatively simple but still incorporates a large 3D context in its decision. During training, the network was fed randomly sampled 60$\times{}$60$\times{}$60 crops in order to stay within the limitations of GPU memory.


\subsection{3D U-Net}

\begin{figure}[tbh]
\centering
\includegraphics[scale=0.6,width=315pt]{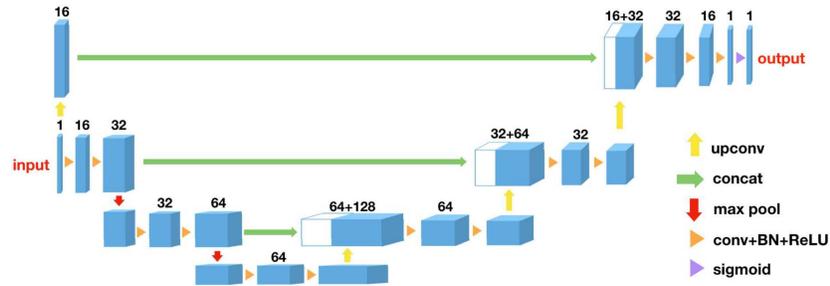}
\caption{3D Superresolution U-Net. The output has twice the input resolution.}\label{fig:3d_unet}
\end{figure}

Like the original 3D U-Net, our network consists of an encoder and a decoder (Figure~\ref{fig:3d_unet}). The encoder contains three convolutional modules, each with two convolutional layers and an increasing number of channels. To avoid introducing misleading information through padding, only valid convolutions are used. Adjacent convolutional modules are connected by a maxpooling layer for downsampling. The decoder part consists of three convolutional modules which are connected by 2$\times{}$ transposed convolutional layers for upsampling. Each upsampled tensor is concatenated with an intermediate output from the encoder part.  
In addition to entering the encoder-decoder path, the input is also directly upsampled with 2$\times{}$ transposed convolution and concatenated with the super-resolution intermediate output of the decoder. At the segmentation output, the channel number is reduced to one and a sigmoid function returns root probability estimates between 0 and 1.

\subsection{Dataset}

The dataset used in this work builds on the synthetic dataset generated by Uzman et al.~\cite{uzman2019learning}. The images in it are generated by combining augmentations of four reconstructed root structures with synthetic soil images which simulate real soil features. 
\paragraph{Combining generated roots with real soil MRI}
To increase the diversity of root structures, we generated 30 random roots of various shapes and root-diameters. 
Furthermore, we used MRI scans of pure soil to capture a larger range of 3D features of soil.
For training, random crops of the generated roots were combined with crops of pure soil and augmented by varying parameters such as the contrast between root and soil. 

For validation, random crops of a different subset of the dataset were used. The final evaluation of the network's performance was done on five real MRI images~\cite{stingaciu2013situ} with human annotations.

\subsection{Loss}

The loss function for training is binary cross-entropy loss averaged among all image voxels. On top of that, we investigated two loss modifications. The first one is applying a higher weight on the root voxels than the soil voxels, forcing the network to focus more on the correct prediction of roots. The reason for trying this is the imbalance of the soil voxel quantity and the root voxel quantity. The second modification is using a don't-care flag to label the root-soil border area, and ignore the voxels in it when calculating the loss. This is done to check the effect of making the learning task easier, because the this area is hard to segment precisely and also not so important.

\subsection{Evaluation}

Since the MRI images contain far more soil voxels than root voxels, the F1-Score is used for evaluation because it is robust against class imbalance. The human annotations are slightly misaligned to the real MRI data, which is a common problem that also occurs in 2D root images~\cite{smith2020segmentation}. Because of that, the Distance Tolerant F1-Score as introduced by Uzman et al.~\cite{uzman2019learning} was used. This way, misalignments between the output and the ground truth will be tolerated if the distance between them is not larger than the value of the distance tolerance. 

\section{Results}

\begin{figure}[h!]
\centering
\includegraphics[scale=0.6,width=315pt]{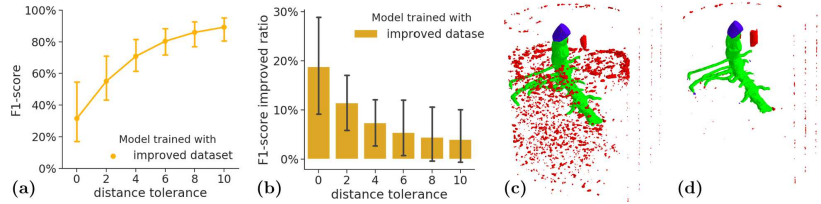}
\caption{Comparison between the models trained with the original dataset and the improved dataset. (a) shows the Distance Tolerant F1-Scores of the model trained with the improved dataset. (b) shows the percentage of improvement of the Distance Tolerant F1-Scores when trained with the improved dataset compared to the original dataset. Error bar indicates the 95\% confidence interval. (c)(d) are example segmentation results of the models trained with the original and improved dataset, respectively. Green, red, blue represent true positives (TPs), false positives (FPs), and false negatives (FNs), respectively.}\label{fig:vis_original_vs_improved_dataset}
\end{figure}

Figure \ref{fig:vis_original_vs_improved_dataset}a shows the Distance Tolerant F1-Scores of the 3D U-Net model trained on the improved dataset, evaluated on the test set with five real MRI scans. For larger tolerances, the score reaches higher values. Figure \ref{fig:vis_original_vs_improved_dataset}b shows the relative improvement of the score, compared to training with the original data set.  The example segmentation results shown in Figure \ref{fig:vis_original_vs_improved_dataset}c, \ref{fig:vis_original_vs_improved_dataset}d demonstrate that  the number of false positives decreases when using the improved dataset. The model trained with the improved dataset reaches an F1-Score of 0.964 on the validation set, and its segmentation results on the test data detect most root branches correctly.  

\begin{figure}[h!]
\centering
\includegraphics[scale=0.6,width=315pt]{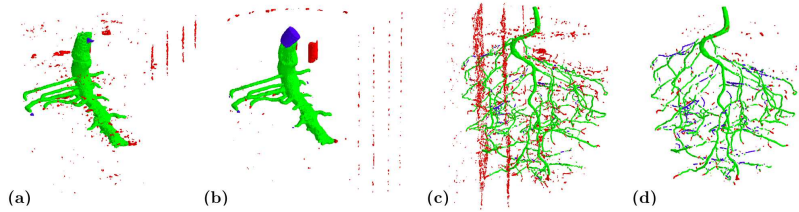}
\caption{Segmentation outputs compared between RefineNet and 3D U-Net on real MRI images. Green, red, blue represent TPs, FPs, and FNs, respectively. (a)(c) are results from RefineNet and (b)(d) are results from 3D U-Net.}\label{fig:compare_with_refinenet}
\end{figure}

Figure~\ref{fig:compare_with_refinenet} compares our 3D U-Net qualitatively to the segmentation results of RefineNet~\cite{uzman2019learning}.  There are fewer false positives in the results of our model, reducing the risk of those false positives being incorrectly extracted as roots by root extraction algorithms. 
However, there are also more false negatives in the 3D U-Net results, which appear as disconnections in the roots. Closing large disconnections can be difficult for the root extraction algorithms. 

\begin{figure}[h!]
\centering
\includegraphics[scale=0.6,width=315pt]{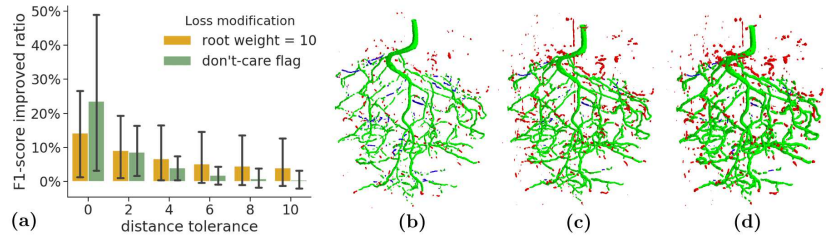}
\caption{Effect of loss modifications. (a) shows the percentage of improvement of the Distance Tolerant F1-Scores of the 2 types of loss modification. Error bar indicates the 95\% confidence interval. (b)(c)(d) shows the example segmentation results of the models trained with the original loss, root weight of 10, and don't-care flag, respectively. Green, red, blue represent TPs, FPs, and FNs, respectively. }\label{fig:later_time_point}
\end{figure}

Furthermore, we investigated two types of loss modifications to deal with the disconnection issue. Here, all compared models are trained with the improved dataset. The first modification is using a higher root weight of 10, which results in improved Distance-Tolerant F1-Scores (Figure~\ref{fig:later_time_point}a) as well as fewer disconnections in the segmented thin roots (Figure~\ref{fig:later_time_point}c). Some root-like false positive predictions are likely roots missed in the human annotations. 
The second loss modification is to ignore the root-soil border area using don't-care flag. This also leads to improved F1-Scores (Figure~\ref{fig:later_time_point}a) but the roots in the segmentation result become visibly thicker which is slightly unrealistic (Figure~\ref{fig:later_time_point}d). The improvement of the F1-Scores is probably largely due to the higher recall because of the thicker root predictions. 

\section{Conclusion}

We investigated semantic segmentation in super-resolution of plant root MRI images using a 3D U-Net model.
When evaluated on the real MRI images, the model can detect most root branches correctly, despite some minor disconnections in thin roots. 
This disconnection problem can be mitigated by applying higher weights on root voxels or ignoring the voxels near the root-soil border when calculating the training loss.


\begin{footnotesize}



\bibliographystyle{myunsrt}
\bibliography{BibFile_3D_UNet.bib}

\end{footnotesize}


\end{document}